\documentclass[conference]{IEEEtran}
\IEEEoverridecommandlockouts
\usepackage{fancyhdr}
\usepackage{cite}
\usepackage{amsmath,amssymb,amsfonts}
\usepackage{algorithmic}
\usepackage{graphicx}
\usepackage{textcomp}
\usepackage{xcolor}
\def\BibTeX{{\rm B\kern-.05em{\sc i\kern-.025em b}\kern-.08em
    T\kern-.1667em\lower.7ex\hbox{E}\kern-.125emX}}

\usepackage{multirow}
\usepackage[utf8]{inputenc}
\usepackage{booktabs}
\usepackage{todonotes}
\usepackage{hyperref}
\usepackage{pifont}
\graphicspath{ {img/} }
\usepackage{blindtext}
\usepackage{cleveref}
\usepackage{fontawesome5}
\usepackage{color, colortbl}
\definecolor{mycyan}{RGB}{141.0, 211.0, 199.0}
\definecolor{myyellow}{RGB}{255.0, 255.0, 179.0}
\definecolor{mylavender}{RGB}{190.0, 186.0, 218.0}
\definecolor{myblue}{RGB}{128.0, 177.0, 211.0}
\usepackage{array}
\newcolumntype{P}[1]{>{\centering\arraybackslash}p{#1}}
\newcolumntype{M}[1]{>{\centering\arraybackslash}m{#1}}

\definecolor{Gray}{gray}{0.9}

\begin{document}

\title{Detecting Inspiring Content on Social Media} 

\author{\IEEEauthorblockN{Oana Ignat}
\IEEEauthorblockA{\textit{University of Michigan} \\
MI, USA \\
oignat@umich.edu}
\and
\IEEEauthorblockN{Y-Lan Boureau}
\IEEEauthorblockA{\textit{Facebook AI} \\
New York City, USA \\
ylan@fb.com}
\and
\IEEEauthorblockN{Jane A. Yu}
\IEEEauthorblockA{\textit{Facebook AI} \\
Menlo Park, USA \\
janeyu@fb.com}
\and
\IEEEauthorblockN{Alon Halevy}
\IEEEauthorblockA{\textit{Facebook AI} \\
Menlo Park, USA \\
ayh@fb.com}
}

\maketitle
\thispagestyle{fancy}

\begin{abstract}
Inspiration moves a person to see new possibilities and transforms the way they perceive their own potential.
Inspiration has received little attention in psychology, and has not been researched before in the NLP community. To the best of our knowledge, this work is the first to study inspiration through machine learning methods. We aim to automatically detect inspiring content from social media data. To this end, we analyze social media posts to tease out what makes a post inspiring and what topics are inspiring. We release a dataset of 5,800 inspiring and 5,800 non-inspiring English-language public post unique ids collected from a dump of Reddit public posts made available by a third party and use linguistic heuristics to automatically detect which social media English-language posts are inspiring.
\end{abstract}


\section{Introduction}

Inspiration is a distinct psychological construct\cite{Thrash2003InspirationAA}, conceptualized by Thrash and Elliot as possessing three core characteristics: evocation (i.e., it is triggered rather than willed), transcendence (i.e., it orients towards things outside of and greater than the self), and approach motivation (i.e., it energizes approach rather than avoidance ~\cite{Thrash2003InspirationAA, Thrash2004InspirationCC, elliot2002approach}). Inspiration has two distinct stages:  one an activation state that is more akin to feeling and emotion, the second an urge to act.  Inspiration is also related to ``other-praising emotions" such as gratitude, elevation, and admiration \cite{algoe2009witnessing}, in that those emotions often result in inspiration. Inspiration, however, is a broader concept and more focused on the resulting increased motivation to act.

\subsection{The appeal of inspiration}
Inspiration is a promising target of study because it is richly associated with a range of positive outcomes, while its absence is described by lay people as something akin to depression\cite{hart1998inspiration}. It can help humans access more creativity\cite{Oleynick2014TheSS}, productivity, and happiness  \cite{thrash2014psychology}, facilitates progress towards goals~\cite{Milyavskaya2012InspiredTG}, and promotes both hedonic (pleasure-oriented)
and eudaimonic (growth-oriented) well-being~\cite{Thrash2010MediatingBT,thrash2010inspiration}. In quantitative analyses of inspiration, Thrash et al. showed that inspiration positively correlates with positive affect, the work-mastery component of 
the need for achievement (but negatively with its competitiveness component), and intrinsic motivation (but negatively with extrinsic motivation) \cite{Thrash2003InspirationAA}. 
They also documented that inspiration appears to cause rather
than simply follow absorption (a state of focused attention on object qualities such as beauty, rather than diffuse arousal), work mastery, creativity, perceived competence, self-esteem, and optimism.
Inspiration can also predict measurable behavioral outcomes such as exploration and purchase \cite{bottger2017customer}, and correlates with outcomes such as higher number
of patents \cite{Thrash2003InspirationAA}. Recall of inspiring content has also been shown to result in significantly more frequent persistence in an unrelated physical endurance task, compared to mere recall of an amusing event \cite{klein2018can}, with people recalling an amusing event being 5 times more likely to give up than those recalling an inspiring one.

\subsection{Scope of this work}
Despite the attractive promise of inspiration, there has been little work on automatically detecting content that is specifically inspiring, rather than merely engaging or positive.
Inspiration requires an ``encounter" \cite{hart1998inspiration}, an event ``in which new or better possibilities are revealed by, or revealed in an evocative stimulus object” \cite{Thrash2004InspirationCC}. 
Our work aims to facilitate such encounters by providing tools for automatic identification of text content likely to be judged inspiring. We focus on inspiration in everyday content as judged by lay people, similar in spirit to early work by Hart who attempted to capture the experience of inspiration in ordinary life \cite{hart1998inspiration}, rather than ``as if it were reserved for the gifted artist, the breakthrough scientist, or the extraordinary mystic."
This motivates our focus on social media content, as well as our use of crowdsourcing, rather than expert labeling.

The contribution of this work is threefold. (1) We introduce the novel task of automatic detection of inspiring content from social media data. To facilitate the task, we offer a novel dataset of inspiring and non-inspiring post ids from Reddit, sourced from a dump of public posts made available by a third party, pushshift.io \cite{baumgartner2020pushshift}. (2) We combine experiments comparing several weak labeling
techniques to retrieve English-language inspirational content with a classifier
trained on human labels to provide a strong baseline to determine if an English-language post is inspiring or not. 
(3) We provide a detailed analysis of data labeled as inspiring to gain insights on which  topics are inspiring, and how they influence the readers.

\section{Related Work}


\subsection{Characterizing inspiring content and its role}
Several works have analyzed characteristics of inspiring content. 
Interviews in the domain of sports have revealed features of coach speeches to athletes that were deemed inspiring, both in terms of content and delivery \cite{smith2018investigating}, for example showing that inspiring speeches provide an illustration of a path towards success or an increase in the sense of belonging, but the work is limited to a sample of young, male athletes.

There have also been some recent papers on inspiring or self-transcendent content (i.e., orienting people towards things outside of or greater than the self) in relation to social media. Oliver et al. \cite{oliver2018self} argues for the importance of studying self-transcendent content to promote well-being and other-oriented connectedness.
Meier et al. showed that benign envy triggered by social media content could lead to inspiration and increased positive affect\cite{meier2018positive}, and that exposure to enhanced positive content on instagram led to higher feelings of inspiration and brief improvements in well-being\cite{meier2020instagram}. This is in contrast with findings of negative consequences when inspiration was not involved\cite{noon2019inspired}.
According to \cite{raney2018profiling}, social media is an important source of inspiring content, especially for younger audiences.
Dale et al. studied 100 videos on YouTube tagged with ``inspiration," and identified a variety of specific features in media eliciting self-transcendent emotion, such as triggers associated with hope (e.g., ``underdog narratives"), appreciation of beauty and excellence (e.g., nature, art, vastness)\cite{dale2017youtube}. Similar triggers were found in “\#inspirational” and “\#meaningful” Tumblr memes\cite{rieger2019daily}, and Facebook posts 
\cite{dale2020self}. 
\cite{dale2017youtube} also reported that 76.5\% of surveyed users had been inspired at some point
by social media content, with 34.9\% reporting this being the case within the past week,
and the most commonly reported inspiring themes being kindness, overcoming obstacles/perseverance, human connection, and gratitude or thankfulness.

\subsection{Quantifying word use in inspiring content}
Ji et al.\ explored textual aspects of inspiring New York Times articles reshared on social media \cite{ji2019spreading,ji2020developing}. They constructed a Self-Transcendent Emotion Dictionary (STED) to detect the presence of 370 English words, word stems, and phrases corresponding to six self-transcendent emotional experiences (awe, admiration, elevation, gratitude, hope, and general inspiration), and validated how it could be used to quantify self-transcendent qualities in text material without resorting to sophisticated natural language processing techniques. However, \cite{ji2020developing} reports that ``the observed effects sizes appear to be relatively small," making counts of STED words an insufficient tool to detect inspiring material, all the more so for social media content where word counts are considerably smaller than in news articles and thus extremely noisy. The STED could nonetheless have been useful as one of the filtering heuristics to select posts over which to train our classifier, however it was not yet available at the time we conducted this work. Another difference consists in the target of the word pattern use: in \cite{ji2019spreading,ji2020developing}, the word patterns apply only to the content itself, rather than also to comments about the content, as we do in this work to capture the fact that inspiration involves elicitation of a response in someone else.

\subsection{Machine learning methods for affective computing}

Despite the fact that inspiration is not, strictly speaking, an emotion, the work closest to ours in terms of the techniques used is in the broader context of emotion recognition (e.g.,~\cite{Strapparava2007SemEval2007T1, Strapparava2008LearningTI}). Emotion recognition is a sub-field of Affective Computing whose goal is to enable machines to understand and emulate affect and emotion~\cite{Picard1997AffectiveCB}. The first benchmark for emotion recognition, Affective Text~\cite{Strapparava2007SemEval2007T1, Bostan2018AnAO} opened the field to several emotion datasets that vary in size, domain, taxonomy and applications: news headlines~\cite{Strapparava2007SemEval2007T1}, tweets~\cite{CrowdFlower, Kharde2016SentimentAO, Mohammad2018SemEval2018T1}, Reddit posts~\cite{Demszky2020GoEmotionsAD}, narrative sequences~\cite{Liu2019DENSAD} and dialog~\cite{Li2017DailyDialogAM, Miller2017ParlAIAD}. One of the key aspects distinguishing our work from the above is that while other works try to detect the emotion of the {\em publisher} of the content, our goal is to detect whether the {\em consumer} of the content would be inspired or not.


To the best of our knowledge, there hasn't been any work using machine learning techniques for textual content, except for Inspirobot~\cite{inspirobot}, which aims to automatically generate inspiring content, but appears more geared towards satire and entertainment than a serious study of inspiration.

\section{Data Collection and Annotation}
\label{data_collection}
In order to develop models for recognizing inspiring content, we need test data annotated with human annotations.
This section describes the details of our dataset.

\subsection{Problem definition}
\label{sec:problem}

The goal of this research is to develop models that can recognize whether a post on social media is likely to inspire someone who reads that post. We focus on text, but recent work on recognizing human intent of images~\cite{DBLP:journals/corr/abs-2011-05558} could form the basis for considering multi-modal content in future work. 

Whether a social media post is inspiring to a user depends on the user and their current state and motivations in life, nevertheless the models we discuss in this paper are not intended to make predictions for specific users. Rather, we aim to predict whether a piece of content would be generally rated as inspiring by human annotators. We rely on a lay understanding of inspiration, following Thrash and Elliot, who confirmed the relative consistency of how inspiration is interpreted by observing measurement invariance through time and populations  \cite{Thrash2003InspirationAA}, and
cited interview research by Hart:
``Despite the
great many shades of meaning the term has in common usage it
appears to represent a clear and consistent event”  \cite{hart1993inspiration,hart1998inspiration}.

\subsection{Data collection}
We annotate two datasets. The first, which we refer to as pR, is taken from a dump of public
posts made available by a third party (pushshift.io Reddit \cite{baumgartner2020pushshift}). The second dataset, which we refer to as S, consists of public social media posts from Facebook. The dataset we make available is based on pR. 

The first challenge in creating the dataset for annotation is that the vast majority of posts on social media are not inspiring. Therefore, in order to ensure that the annotators are exposed to a sizeable amount of inspiring posts, we filter the data using the following heuristics: (1) public posts with at least one comment that contains the substrings ``inspir" or ``uplift" (pR \& S), (2) public posts that authors mark as ``feeling inspired" or ``feeling up" (S), (3) public posts that are shared at least 10 times (S), (4) public posts from the subreddits that contain the substrings ``inspir" or ``uplift" (pR), and (5) comments to the following four questions from the ``AskReddit" subreddit:  ``When was the last time you felt inspired?", ``Who or what inspired you?", ``Who inspired you and how?", ``What is the most inspiring thing you have ever seen or heard?"  (pR).
As control, we also collect random posts: (1) posts with no comment that contains the substrings ``inspir" or ``uplift" (pR \& S), (2) posts from random subreddits that do not contain the substrings ``inspir" or ``uplift" (pR).

\subsection{Data pre-processing}
We filter out posts that are not in English using the language classifier from fastText\cite{bojanowski2016enriching}. We remove links and double spaces and we filter out posts that contain less than 10 words or more than 200 words.
We remove posts that contain offensive language or profanity using an SVM model\cite{Cortes2004SupportVectorN} trained on two datasets of hate speech in social media~\cite{Schofield2017IdentifyingHS, KaggleToxicCommentChallenge}. This basic process of removing offensive language could be improved, for example by using careful human labeling to decide what should not appear in the dataset. However, there is not yet a fully agreed upon method, and best practices for diverse and inclusive data collection are still very much a matter of active research \cite{Davidson2017AutomatedHS, Sap2019TheRO, Davidson2019RacialBI, Mozafari2020HateSD, Zhou2021ChallengesIA}.

For data analysis, we further pre-process the data: we remove stop words and specialized words (\textit{subscribe}, \textit{comment}, \textit{like}, \textit{follow}, \textit{link}), special characters and digits, hashtags, punctuation, and emojis, whose prevalence in the dataset was not high enough for analysis. We also convert the words to lowercase and lemmatize them.  

\subsection{Data annotation}
\begin{figure}
\centering
\includegraphics[width=\linewidth]{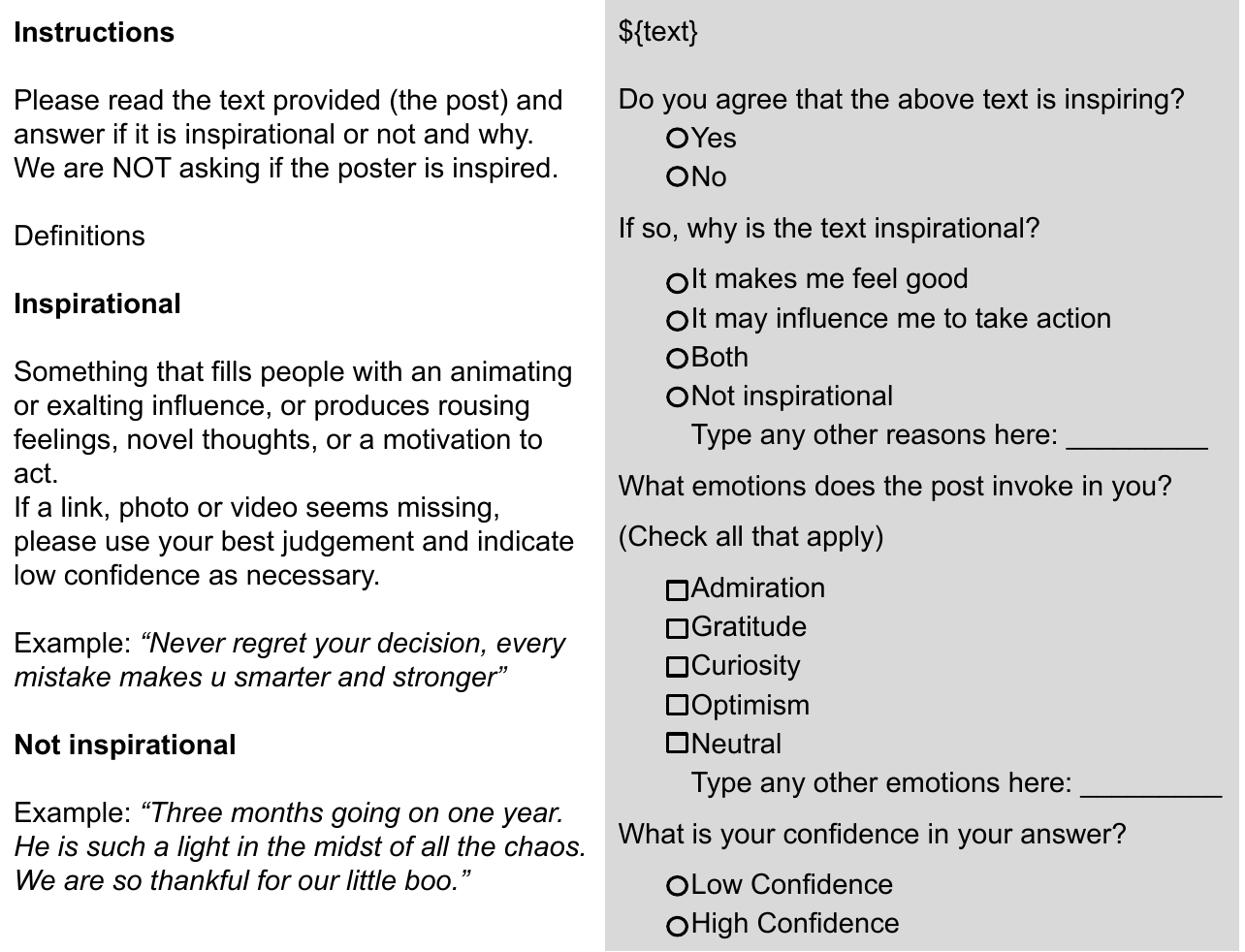}
\caption{User Interface for the Amazon Mechanical Turk annotation platform.}
\label{fig:AMT_GUI}
\end{figure}

The resulting posts are annotated by crowdsourced workers  to determine: (1) whether the post is inspiring or not; (2) if the post is inspiring, what influence it has on the reader; (3) what emotions it evokes; (4) the annotator's confidence in the answer.
The instructions and interface for annotation are shown in Figure~\ref{fig:AMT_GUI}. We provide the annotators with a short definition of what inspiration is, one example of an inspiring post and one example of a non-inspiring post. The two example posts from the guidelines are selected by the paper authors, among the filtered posts.

Each post is labeled by three different annotators.
We compute the agreement between the annotators using Fleiss Kappa measure \cite{fleiss1973equivalence} and we obtain 0.26, indicating fair agreement yet showing that the task remains subjective.

Due to the challenge of finding inspiring posts at scale, our annotation proceeded in several phases. In the first phase, we 
annotated 1000 pR posts: 500 from the subreddits that contain the substrings ``inspir" or ``uplift" and 500 from random subreddits. We find that only 167 posts are labeled as inspiring by at least two annotators, and most of these posts are of a particular genre of inspiring quotes. We also find very few false positive posts (9 /500), which suggests that non-inspiring posts are much easier to collect. Therefore, in the next annotation phases, we focus on collecting inspiring posts.

In order to obtain a bigger fraction of inspiring posts, we changed the filtering heuristic in the second phase to consider the content of comments on posts. Specifically, we collect posts with at least one comment that contains the substrings ``inspir" or ``uplift" and responses to posts asking questions that contain the substring ``inspir", from the ``AskReddit" subreddit (pR): ``When was the last time you felt inspired?", ``Who or what inspired you?", ``Who inspired you and how?", ``What is the most inspiring thing you have ever seen or heard?". These posts turn out to be much more diverse and do not focus on inspiring quotes. 
From a total number of 3,415 annotated posts, 1,225 posts are labeled as inspiring by at least two annotators - considerably more than the first phase. 

In the third phase we use the previously annotated posts to fine-tune a pre-trained RoBERTa model~\cite{Liu2019RoBERTaAR} to the task of predicting inspiring content.  The trained classifier is then used as additional filtering to select the
posts to be annotated from the ones that fit the second heuristic.
A third set of posts selected in that process is
annotated by workers.
From 3,210 annotated posts, 3,054 are labeled as inspiring by at least two annotators. 
This final selection pipeline thus seems to
be predictive of subsequent human annotation as "inspiring."\footnote{Note that despite our efforts to obtain a wide variety of inspiring posts, we do not have a measure of the inspiring posts that we missed, only results showing that the posts we select are largely deemed inspiring.}

Non-inspiring posts are posts collected from random subreddits and posts that match the heuristics yet were not labeled inspiring.
A post is labeled as non-inspiring if the trained classifier labeled it as non-inspiring or if no annotator labeled it as inspiring.
After balancing the number of inspiring and non-inspiring posts, we obtain 11,592 total posts: 5,796 inspiring and 5,796 non-inspiring.  A post is inspiring if at least one annotator labeled it as inspiring. The inspiring post is assigned an agreement score corresponding to the number of annotators who labeled it as inspiring (3/3 is the maximum agreement score and 1/3 is the minimum).
Note that the agreement score of the post is different from the confidence of the annotator, which can be low or high. We find that the annotators are highly confident in their answers 91\% of the time, and thus we decided to not use the confidence of the annotators in our experiments.
We show the annotation results in \Cref{tab:annotation_results}.

\begin{table}
\centering
\setlength{\tabcolsep}{0.5em} 
{\renewcommand{\arraystretch}{1.3}
\begin{tabular}{c |M{1.5cm}|M{1.5cm}|M{1.5cm}}
\multirow{ 2}{*}{\sc \# non-inspiring} & \multicolumn{3}{c}{\sc \# inspiring} \\
  & 1/3 & 2/3 & 3/3 \\
  \midrule
  5,796 & 1,517 & 1,936 & 2,343 \\
      \end{tabular}
      }
      \newline
    \caption{Number of human annotated posts: inspiring and non-inspiring, grouped by agreement score.}
    \label{tab:annotation_results}
\end{table}

\section{Classification of inspiring posts}
We now describe our baseline classification methods for the task of detecting inspiring posts on social media.
We experimented with 
 fastText~\cite{joulin2017bag}, and fine-tuned pre-trained language models BERT~\cite{Devlin2019BERTPO} and RoBERTa~\cite{Liu2019RoBERTaAR}. In each case, we fine-tuned the model for our task using our annotated training data.  We balance the training data to have an equal number of inspiring and non-inspiring posts. The statistics for this experimental split is shown in~\Cref{tab:train-test-eval-split}. In the experiments, we used early stopping, 5 training epochs and a batch size of 64. More training epochs (10 and 20) did not improve the results. 

The results, shown in \Cref{tab:all_results}, show that the model has a high accuracy when distinguishing inspiring from non-inspiring posts. In the experiments we train the model on each dataset separately and the best performing model is RoBERTa. 
In addition, we tried a transfer learning experiment where we trained the classifier on the dataset S and tested it on the data in pR (last line in \Cref{tab:all_results}). 
This results in lower performance than training specifically on pR, as expected for data from a different distribution, but there is still some transfer of performance.
We hypothesize that the transfer learning was not stronger because of the different nature of the posts in the two datasets: while the posts in S tend to be more personal short stories or announcements, the inspiring posts in pR tend to be more general and broad, such as news, discussions, and asking or answering questions.
 
During error analysis, we find that the most common errors our best model makes are false negatives: posts that are collected as inspiring, but labeled by the model as non-inspiring. We find that in most cases, the posts that confuse the model are the ones annotated as inspiring by one annotator.

\begin{table}
\centering
\setlength{\tabcolsep}{0.5em} 
{\renewcommand{\arraystretch}{1.3}
\begin{tabular}{@{\extracolsep{7pt}}l c r c c@{}}
  \toprule 
     &
        \multicolumn{2}{c}{\textbf{S}} &
      \multicolumn{2}{c}{\textbf{pR}} \\
      \cline{2-3} \cline{4-5}
  \multicolumn{1}{l}{} & inspiring &  \multicolumn{1}{c}{non-inspiring} &  
  \multicolumn{1}{c}{inspiring} &  \multicolumn{1}{c}{non-inspiring}\\
  \midrule
  Train & 642,517 & 642,517 & 5,216 & 5,216 \\ 
  Test & 71,390 & 71,390 & 580 & 580 \\
\bottomrule
      \end{tabular}
      }
      \newline
    \caption{Number of posts for the experimental data split.}
    \label{tab:train-test-eval-split}
\end{table}

\begin{table}
\centering
\setlength{\tabcolsep}{0.6em} 
{\renewcommand{\arraystretch}{1.3}
\begin{tabular}{@{\extracolsep{7pt}}l c r c c@{}}
  \toprule 
     &
    \multicolumn{2}{c}{\textbf{S}} &
      \multicolumn{2}{c}{\textbf{pR}} 
      \\
      \cline{2-3} \cline{4-5} 
  \multicolumn{1}{l}{Method} & Acc. &  \multicolumn{1}{c}{F1} &  
  \multicolumn{1}{c}{Acc.} & \multicolumn{1}{c}{F1} \\
  \midrule
  fastText & 74.66 & 74.58  & 76.20 & 76.20 \\ 
  BERT & 75.18 & 76.22 & 83.53 & 84.23 \\
  RoBERTa & 79.81 & 80.73 & \textbf{87.24} & \textbf{88.08}\\
  RoBERTa* & 79.81 & 80.73 & 67.93 & 71.73\\
\bottomrule
      \end{tabular}
      }
      \newline
    \caption{Results from baselines and main method on S and pR test data. * - model was fine-tuned on S data. S is weakly labeled and pR is human annotated.}
    \label{tab:all_results}
\end{table}

\section{Analyzing inspiring posts}
An important goal of our research is to gain insights into what makes a social media post inspiring. Extensive analysis of our data through topics and sentiment analyses confirm the following about inspiring content on social media: \textit{inspiration is a complex state which can contain both positive and negative emotions, and can be found across a diverse range of every-day moments and activities.} 

We analyse two different types of inspiring posts: posts that \textit{make people feel good} by changing their mood or their perspective (e.g., feel gratitude, admiration, curiosity), and posts that \textit{make people act} on their thoughts and wishes (e.g., start a new hobby or a new work routine). \Cref{tab:effect_stats} shows the split among these categories from our annotated data. \Cref{tab:effect_eg} shows example posts from each category.  Our annotation also allowed workers to add other motives for why they find the post inspiring. Examples from the ``other" category include: ``its practical life", ``its real truth", and comments that relate the post to their previous life experience, such as ``I was the second fastest sprinter female in my school", or ``people are willing to go through so much for their animals". 

\begin{table}
\centering
\setlength{\tabcolsep}{0.5em} 
{\renewcommand{\arraystretch}{1.3}
\begin{tabular}{l|M{1.5cm}|M{1.5cm}|M{1.5cm}|M{1.5cm}}
  \multicolumn{5}{c}{\sc \# inspiring posts (5,796 in total)} \\
   \midrule
    & \sc motivation to act  & \sc feel good  &  \sc no effect & \sc other \\ 
    \midrule
    $\geq$ 1 & 3,333 & 4,382 & 3,470 & 564 \\ 
    $\geq$ 2 & 1,052 & 1,711 & 1,476 &  13 \\
    $\geq$ 3 & 149   & 317   & 20    &   0 \\
      \end{tabular}
      }
      \newline
    \caption{Number of inspiring posts labeled with their effects on the readers by at least one, two and three (all) annotators.}
    \label{tab:effect_stats}
\end{table}

\begin{table*}[!ht]
    \begin{tabular}{M{0.47\textwidth} | M{0.47\textwidth}}
        \sc motivation to act  & \sc feel good\\
        \midrule
        \textit{``This is very inspiring to me, as I will be returning to college soon to begin schooling in Physics. Being schizophrenic can (and will) make you think you won't get far in life, and its very easy to fall into a non-productive spiral really quick. People like John Forbes Nash Jr. and this fellow can really inspire without even meaning to, and that gives me a lot of hope for my future. Good Post."} & \textit{``I say follow your heart. Play what you're inspired to play. If you're inspired to polish a piece, work on that. If you're inspired to just relax and play pieces you already know, do that. The most productive practicing happens when your mind is most concentrated, and when you're inspired by something, your mind is automatically attracted to that thing. I think it's best to take advantage to that phenomenon."}\\[0.5cm]
        \hline
        \textit{``Just saying you inspired my dad to become a bodybuilder, and inspire me to work out and stay fit. Don't let anyone get you down, you're the terminator! Your a good example about you can do anything if you set your mind to it!"} & \textit{``As simple as this sounds, I'm happy that I have a roof to live under, food to eat, and people that care about me. We all take that for granted until its all gone."}\\
        \hline
        \textit{``Do something to help or enrich the lives of others. Could be volunteering at a youth/vulnerable peoples centre, getting involved in a local art project, or making a set of instructional videos on youtube about something you know a lot about. If you inspire just a single person you're still making the world that little bit better."} & \textit{``Motivation is key.Some mornings I look out, it's cold raining wet and miserable, I just pop in my headphones and play some uplifting music, stuff to get you pumped up.Failing that, run with others. That way you have the motivation of not letting others down. Join a crew so if you cancel it's not yourself you're letting down, but the crew."}\\
    \end{tabular}
      \newline
    \caption{Examples of inspiring posts from pR with different influence on the reader.}
    \label{tab:effect_eg}
\end{table*}

From our data observations, the inspiring posts can also be categorized as \textit{general} or \textit{personal}. Posts like inspiring news or quotes apply to most people and circumstances as they promote popular human values such as gratitude, determination or kindness (e.g., \textit{``Don't be content everyday to do no wrong, be prepared every day to do good" @6:50 in that video, definitely advice to live by"}). These are different from the more personal posts such as someone sharing how they overcame a personal struggle (e.g., \textit{``Over the past year I've lost some weight, but have been feeling pretty down and hate how most of my clothes look. You have inspired me to donate what I hate and go get some damn clothes that fit. Thank you!"}).

\subsection{Topic analysis}
Inspiration is present across multiple topics: \textit{spiritual} (e.g.,  religion, mindfulness), \textit{physical} (e.g., hobbies, travel) and \textit{artistic} (e.g., dance, painting). In what follows, we analyze the distribution of these topics.

\medskip
\noindent
{\bf TF-IDF score:}
 We compute the tf-idf~\cite{Ramos2003UsingTT} scores of all the words in a document composed of all the inspiring and non-inspiring posts. We show the top 20 words from inspiring posts, sorted by their tf-idf score in \Cref{tab:stats_data_analysis}.

\medskip
\noindent
{\bf Word frequency:}
Table~\ref{tab:stats_data_analysis}  shows the words that are most frequent in the inspiring posts, compared to the non-inspiring ones, the top 20 frequent bi-grams and tri-grams and the top frequent hashtags in the inspiring posts. The hashtags act like keywords, summarizing the content of the S and pR posts. The words can be grouped into 4 main categories: (a) names of famous figures (e.g., philosophers, authors, athletes, activists, scientists); (b) human values; (c) spirituality; (d) art and beauty.

\begin{figure}
\centering
\includegraphics[width=\linewidth]{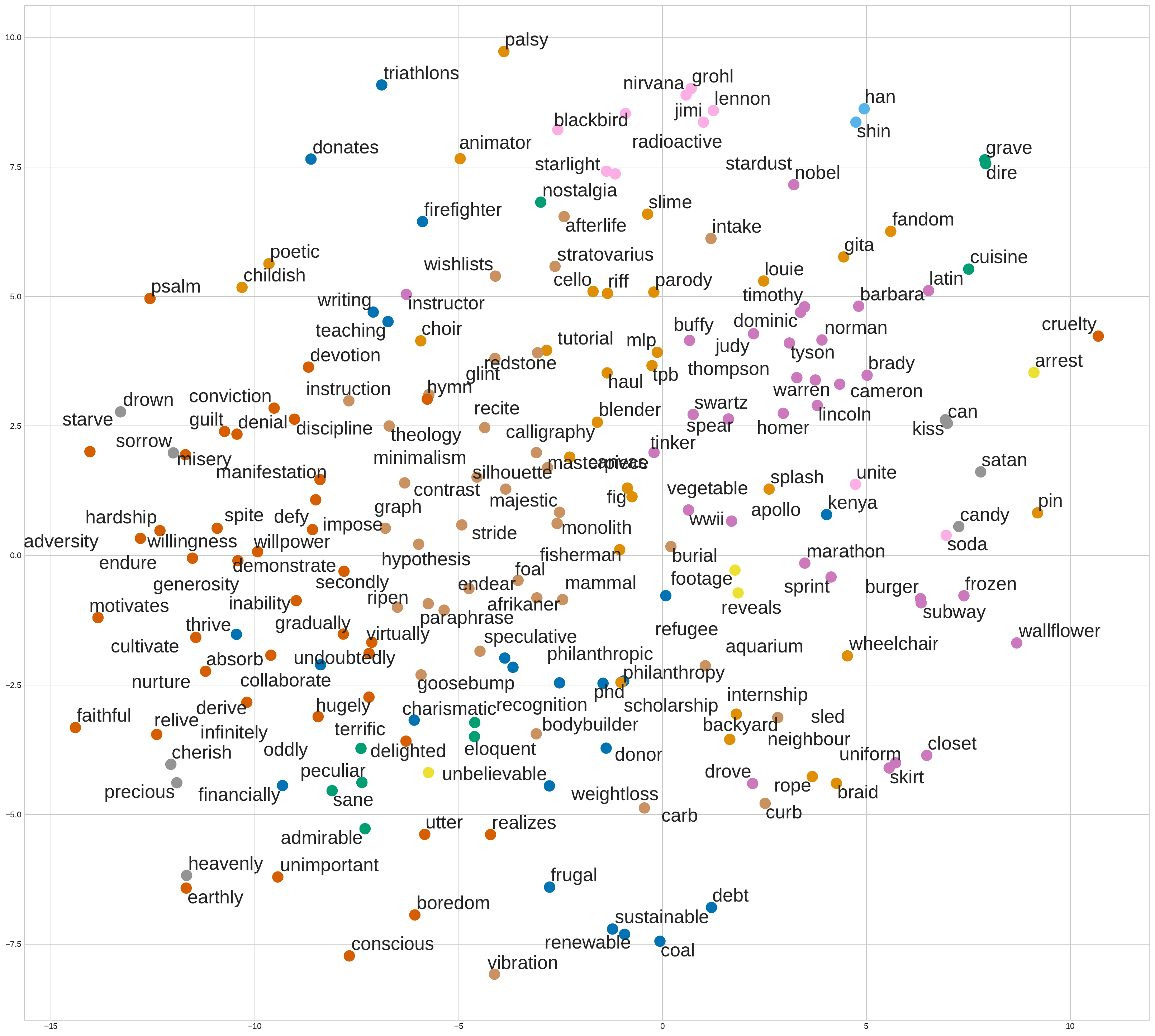}
\caption{K-means\cite{Hartigan1979AKC} clustering of most important (sorted by their tf-idf score) 200 word embeddings from inspiring pR posts.}
\label{fig:clustering_reddit}
\end{figure}

\begin{figure}
\centering
\includegraphics[width=1\linewidth]{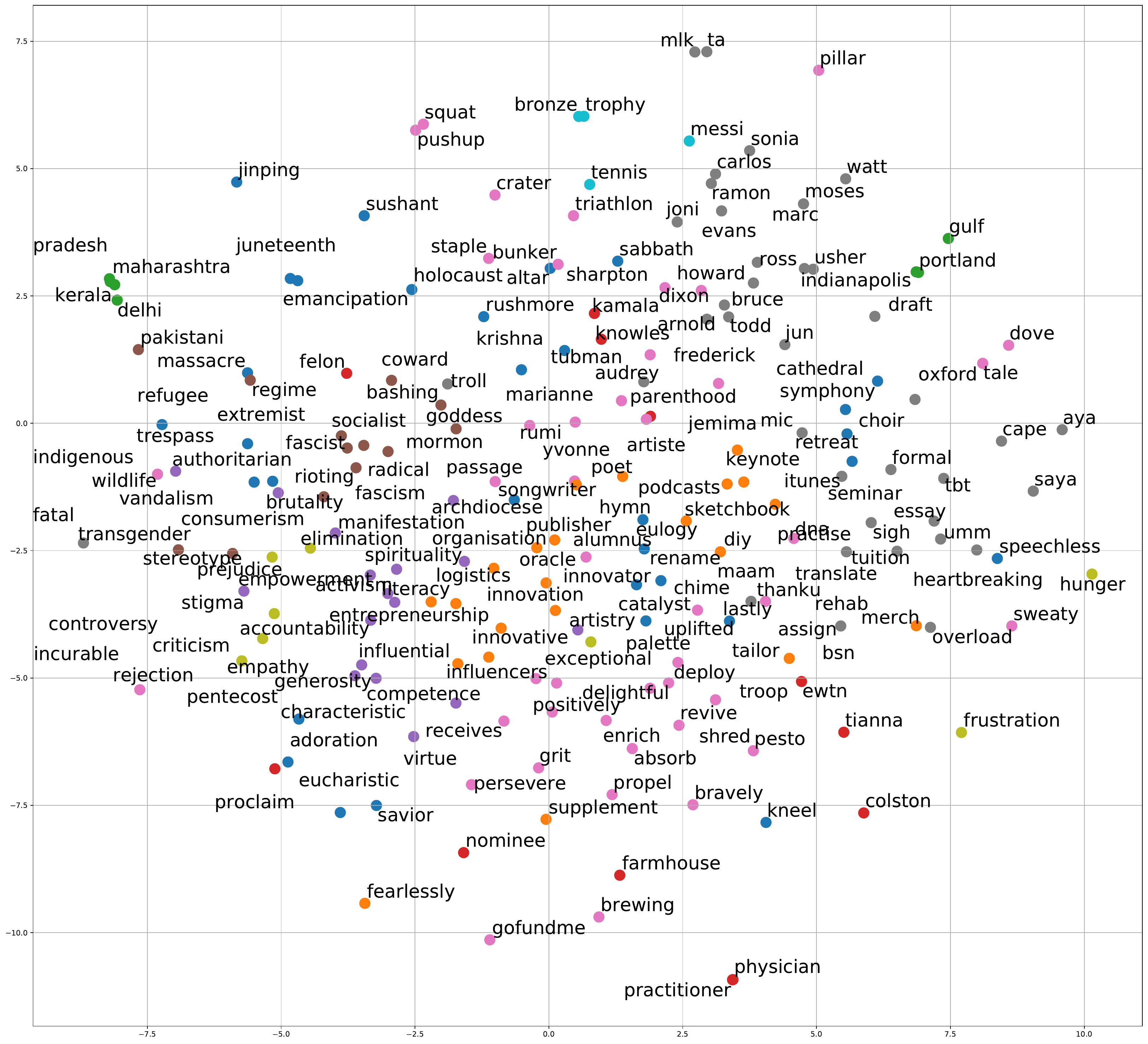}
\caption{K-means\cite{Hartigan1979AKC} clustering of most important (sorted by their tf-idf score) 200 word embeddings from inspiring S posts.}
\label{fig:clustering_fb}
\end{figure}

\medskip
\noindent
{\bf Clustering topics:}
We represent the words using GloVe~\cite{Pennington2014GloveGV} embeddings, the 200 dimensional model trained on two billion Tweets. We apply k-means~\cite{Hartigan1979AKC} clustering on the most important (sorted by their tf-idf score) 200 words from inspiring posts. Using the elbow method, we determine that the best number of clusters is 10. After applying t-distributed Stochastic Neighbor Embedding (t-SNE)~\cite{Maaten2008VisualizingDU}, the two dimensional plots can be seen in \cref{fig:clustering_reddit} and \cref{fig:clustering_fb}.

From the clusters, we observe words related to spiritual domain: \textit{heavenly, faithful, cherish, conscious, psalm, hymn}; words related to human values: \textit{generosity, philanthropy, willingness, willpower, devotion, charismatic, defy, minimalism}; and words related to artistic or creative pursuits: \textit{poetic, writing, cello, recite, teaching, cuisine}

From the clusters, we can also observe groups of words that express positive and negative emotions: \textit{precious, thrive, generosity, willingness, poetic, philanthropic, majestic, heavenly, cherish, motivates, delighted, scholarship, phd, teaching, firefighter} vs. \textit{misery, adversity, spite, cruelty, arrest, defy, sorrow, starve, endure, WWII, burial, refugee, debt, palsy}. This reflects the fact that inspiration is often found in circumstances where someone rises above adversity.

Examples of these pairs can be seen in posts about overcoming difficulties, which is a popular theme in inspiring posts (e.g., \textit{``That a kid with a younger brother with cerebral palsy runs triathlons with his brother attached to him"}, \textit{``Man, I gotta start running again! The fact that your post has the picture of Emi that I was inspired to draw after beating her arc has me even more motivated!"}, \textit{``96 Year-Old Latino Man Graduates with Top Honor, Survives WWII, and Now the Pandemic"})

\subsection{Emotion and Sentiment Detection}

We also analyze the sentiment and emotions expressed in inspiring posts using automatic classification models and human annotation.

\begin{figure}
    \centering
    \includegraphics[width=0.7\linewidth]{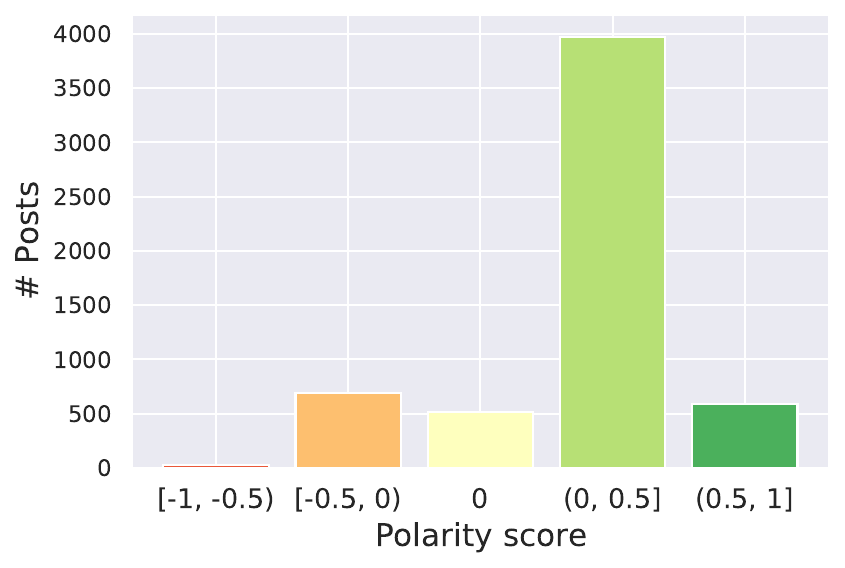}
    \caption{The polarity of all the inspiring posts, grouped into bins by sentiment score: from negative, to neutral and positive sentiment.}
    \label{fig:polarity}
\end{figure}

\paragraph{Sentiment Analysis}
We measure the sentiment of the inspiring posts using polarity\footnote{Polarity is measured using https://textblob.readthedocs.io/en/dev/.}: a score between -1 and 1 which corresponds to how negative or how positive the sentiment of the post is. The results can be seen in \Cref{fig:polarity}. We observe that the inspiring posts with positive sentiment are the most frequent. 

\begin{figure}
    \centering
    \includegraphics[width=\linewidth]{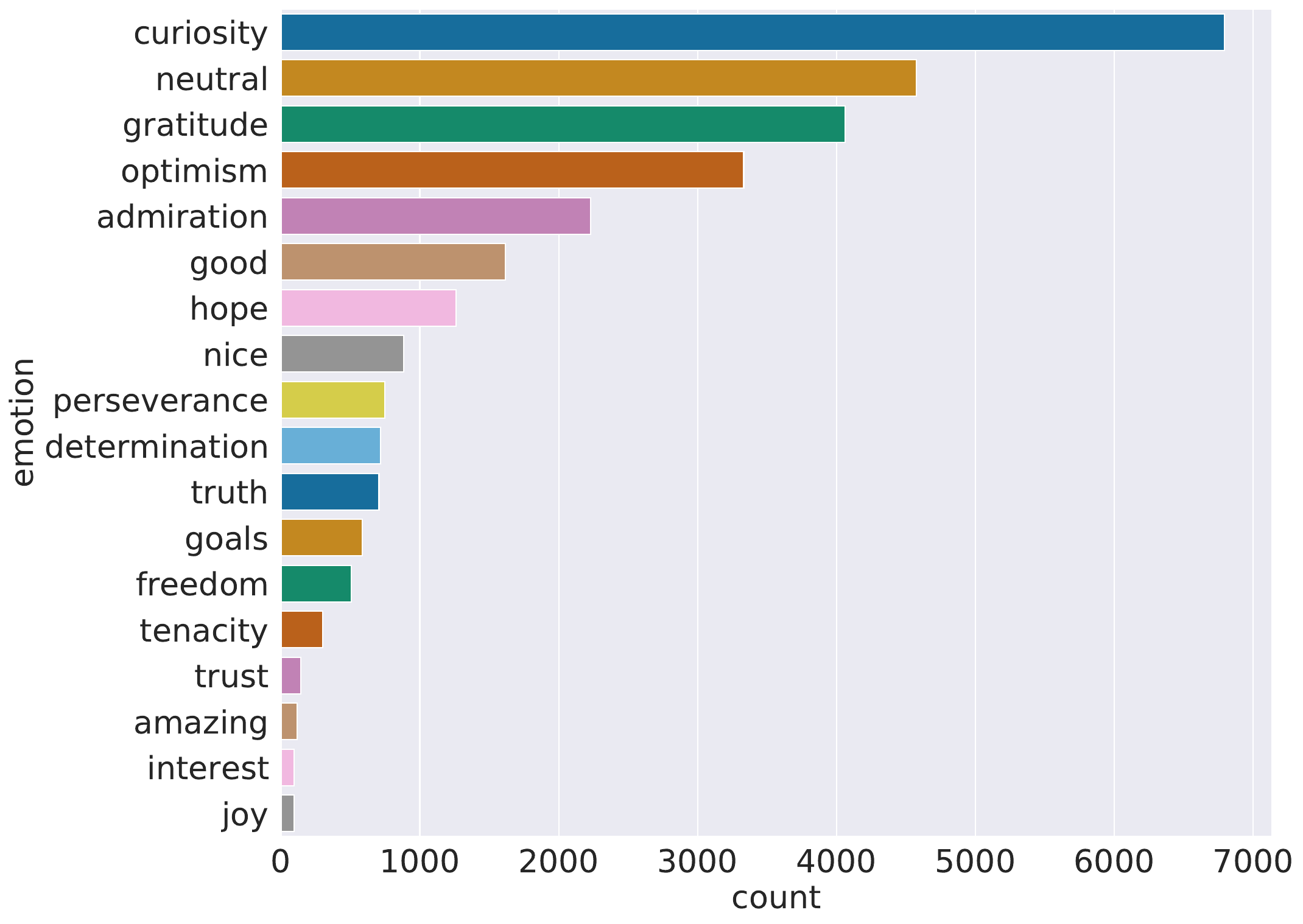}
    \caption{Most frequent emotions labeled by the Amazon Mechanical Turk annotators in the inspiring posts: curiosity, neutral, gratitude, optimism and admiration are solicited in the annotation task, while the others are not.}
    \label{fig:emotions_annotated}
\end{figure}

\paragraph{Emotions}
We also asked annotators to select emotions they felt while reading the posts, among a set of candidates (admiration, gratitude, curiosity, optimism and neutral) that looked likely from previous research (e.g., \cite{algoe2009witnessing}). The annotators also have the option to type in any other emotion they consider relevant (\Cref{fig:emotions_annotated}).

\begin{figure}
    \centering
    \includegraphics[width=0.9\linewidth]{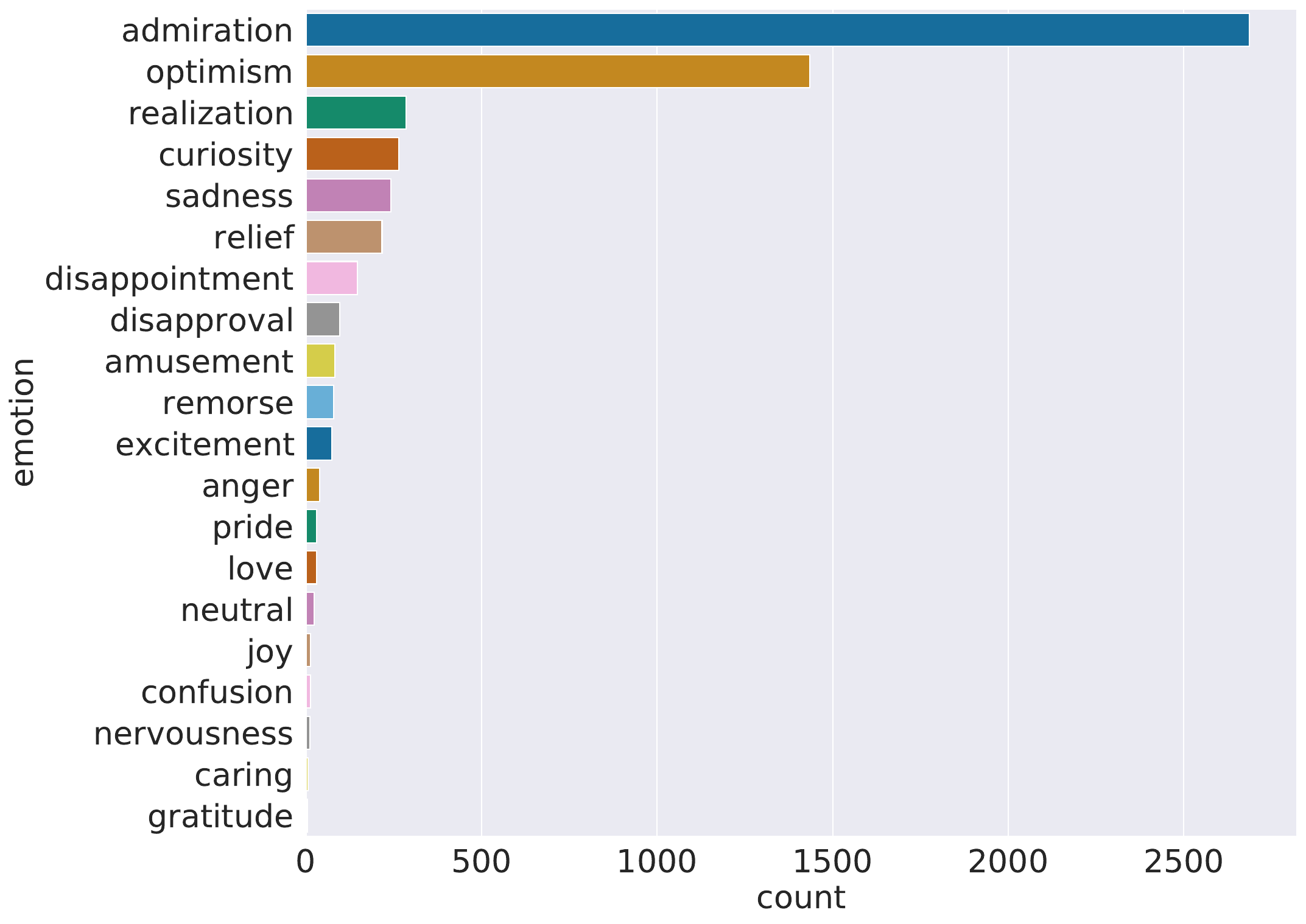}
    \caption{Emotions predicted by fine-tuned RoBERTa\cite{Liu2019RoBERTaAR} in inspiring pR posts.}
    \label{fig:emotions_predicted}
\end{figure}

For the emotion classifier, we use the GoEmotions dataset\cite{Demszky2020GoEmotionsAD}, the largest manually annotated dataset of 58k English Reddit comments, labeled for 27 emotion categories or Neutral.
We fine-tune the RoBERTa\cite{Liu2019RoBERTaAR} multi-class classification model on the GoEmotions dataset and apply the model to our pR posts. Admiration, optimism, sadness and curiosity are selected frequently by the emotion classifier, although gratitude isn't (\Cref{fig:emotions_predicted}).

\begin{figure}
    \centering
    \includegraphics[width=0.9\linewidth]{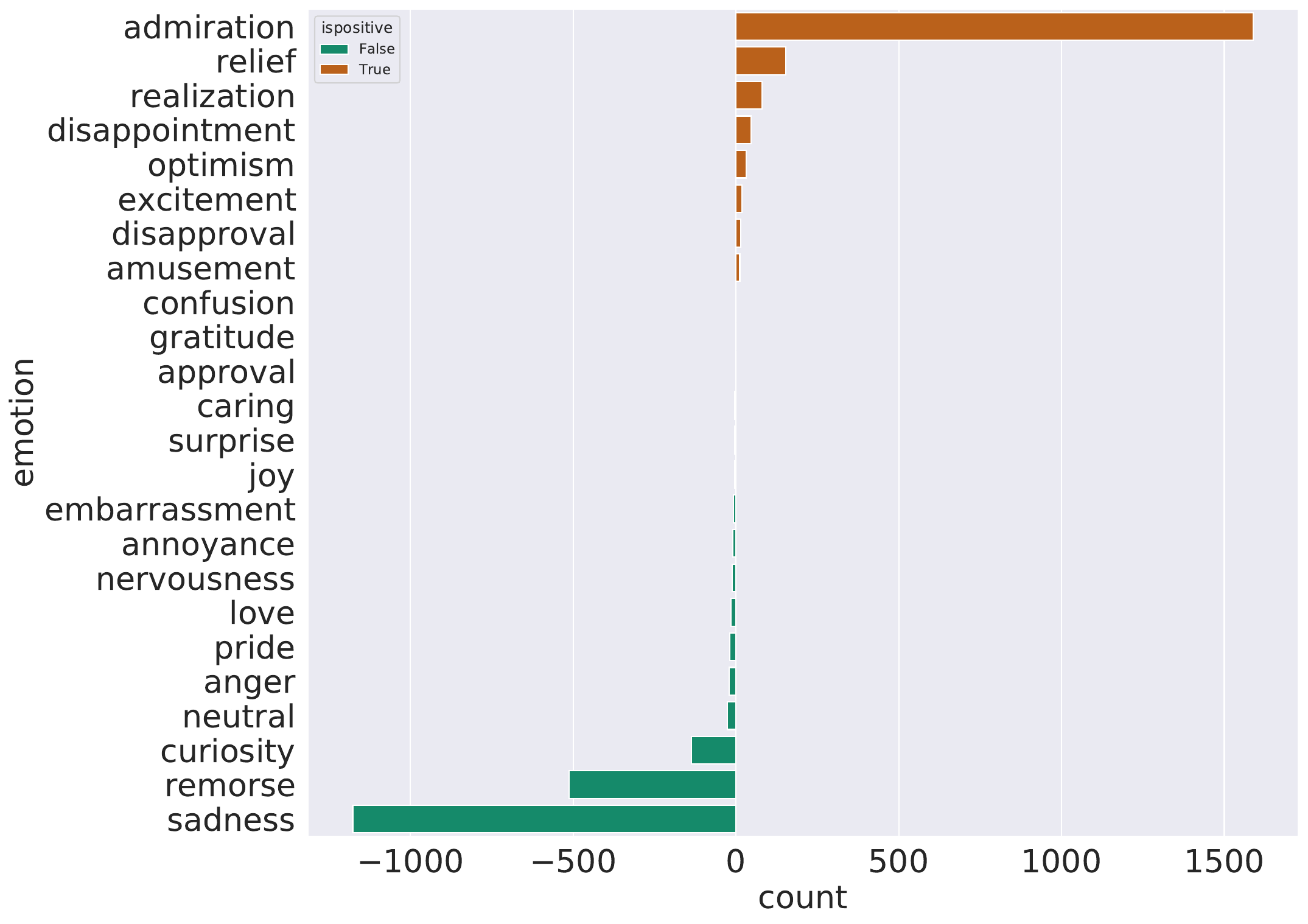}
    \caption{Difference in frequency between emotions detected by RoBERTa\cite{Liu2019RoBERTaAR} model in inspiring and non-inspiring pR posts. Red: emotions that are more prevalent in inspiring posts. Green: emotions that are more prevalent in non-inspiring posts.}
    \label{fig:emotions_comparison}
\end{figure}

Figure \ref{fig:emotions_comparison} shows the difference in frequency of the predicted emotions in inspiring and non-inspiring pR posts. As can be seen, \textit{admiration} and \textit{relief} are the most frequent emotions in inspiring posts, compared to the non-inspiring posts, while \textit{sadness} and \textit{remorse} are more present in the non-inspiring ones. Note that many non-inspiring posts were selected among posts that matched our heuristics, so may not be representative of a purely random set of negative examples.



\begin{table*}[htbp]
\centering
\resizebox{0.85\linewidth}{!}{%
\begin{tabular}{|p{0.45\textwidth} | p{0.45\textwidth}|}
    \toprule
      \hfil \textbf{S} & \hfil \textbf{pR} \\
       \toprule 
      \multicolumn{2}{|p{0.9\textwidth}|}{\centering \textbf{Top words with highest tf-idf scores}}\\
      \toprule
        \colorbox{mycyan}{moses, kamala, jemima};\colorbox{mylavender}{savior, positive, generosity, brutality,} \colorbox{mylavender}{radical};\colorbox{myyellow}{adoration, kneel};
        \colorbox{myblue}{poet, tailor, diy, choir}
        &
       \colorbox{mycyan}{lincoln, lennon, swartz, thompson};\colorbox{mylavender}{guilt, hardship, conscious} \colorbox{mylavender}{nurture, marathon, teaching};\colorbox{myyellow}{choir, grave, psalm}; \colorbox{myblue}{poetic, backyard}
        \\[0.1cm]
      \toprule
      \multicolumn{2}{|p{0.9\textwidth}|}{\centering \textbf{Top words more frequent in inspiring than in other posts}}\\
      \toprule
       \colorbox{mylavender}{share, people, time, make, work, world, support, thank, help},
        \colorbox{myyellow}{feel, journey, live},
        \colorbox{myblue}{story, video, watch}
        &
      \colorbox{mylavender}{hope, amaze, love, great, thank, learn, world, others};\colorbox{myyellow}{feel, dream};
        \colorbox{myblue}{music, beautiful, write}
      \\[0.1cm]
      \toprule
      \multicolumn{2}{|p{0.9\textwidth}|}{\centering \textbf{Most frequent bi-grams}}\\
      \toprule
      \colorbox{mylavender}{black life, life matter, father day, family friend, thank everyone,} \colorbox{mylavender}{happy birthday, hard work, feel free};
     \colorbox{myyellow}{god bless, thank god} &
     \colorbox{mycyan}{david brin, carl sagan};\colorbox{mylavender}{awe inspire, uplift series, thank much} \colorbox{mylavender}{make happy, feel well, lose weight, inspire work, inspire hope} \\[0.1cm]
      \toprule
      \multicolumn{2}{|p{0.9\textwidth}|}{\centering \textbf{Most frequent tri-grams}}\\
      \toprule
      \colorbox{mycyan}{sushant singh rajput};\colorbox{mylavender}{black life matter, happy father day}, \colorbox{mylavender}{stand chance win, wish happy birthday, weight loss journey},
      \colorbox{mylavender}{dream come true, thank family friend};\colorbox{myblue}{new york city}
      &
      \colorbox{mycyan}{david brin uplift, martin luther king};
      \colorbox{mylavender}{keep good work, make feel well} \colorbox{mylavender}{want say thank, inspire many people, rise uplift war} \\[0.1cm]
       \toprule
      \multicolumn{2}{|p{0.9\textwidth}|}{\centering \textbf{Most frequent hashtags}}\\
      \toprule
      \colorbox{mycyan}{\#georgefloyd};
    \colorbox{mylavender}{\#blacklivesmatter, \#love, \#motivation, \#mentalhealth,} \colorbox{mylavender}{\#fathersday, \#success, \#goals, \#giveaway, \#fitness, \#contest, \#grateful,}
    \colorbox{mylavender}{\#womensupportingwomen, \#entrepreneur, \#staysafe};
     \colorbox{myyellow}{\#blessed, \#catholic, \#transformation, \#life}
     & 
    \colorbox{mylavender}{\#thursdaytruthful, \#standup, \#guidelines, \#wecanbe, \#thelift};
    \colorbox{myyellow}{\#balance, \#inrecovery};
    \colorbox{myblue}{\#art, \#artist, \#artistdiscovery}
    \\[0.1cm]
      \bottomrule
      \multicolumn{2}{|p{0.9\textwidth}|}{\centering \textbf{
    \colorbox{mycyan}{Personalities};
    \colorbox{mylavender}{Human values};
     \colorbox{myyellow}{Spirituality};
     \colorbox{myblue}{Art and beauty}}}
    \\
      \bottomrule
      
\end{tabular}%
}

\caption{Statistics from inspiring posts in S and pR data. The data is lemmatized.}
\label{tab:stats_data_analysis}
\end{table*}

\section{Concluding remarks}
In this paper, we introduced the novel task of identifying inspiring posts in social media data. We performed extensive analyses on Reddit (pR) data and other social-media data in order to gain insight into what topics are inspiring, what words are specific to inspiration and what emotions the inspiring posts transmit to readers. To facilitate research in that domain, we release a new dataset consisting of 5,796 inspiring and 5,796 non-inspiring post ids collected from pR. We also annotate these posts with the effect they have on the reader and the emotions they transmit. The dataset is publicly available at \url{https://github.com/MichiganNLP/Reddit-Detect-Inspiration}.

We have already noted in the introduction the numerous benefits of inspiration as a promoter of wellness, creativity, and motivation. Another potential benefit for inspiring social media content is that it might crowd out less desirable content.
A lot of attention has focused on the AI challenges in mitigating  negative aspects of social media, due to their potential role in spreading misinformation, hate speech and other forms of policy violating content~\cite{DBLP:journals/corr/abs-2009-10311}. An indirect method for mitigating such harms is for social network recommendation systems to find more positive and inspiring content to show to users. This paper takes a first step towards identifying inspiring content that can be shown to users. 

Avenues for future work include extending these methods to identify multi-modal inspiring content, analyzing in more detail the effects of inspiring content (including unintended consequences) on users, and analyzing the style of inspiring posts. More powerful classifiers based on larger language models could also be developed to detect inspiring content directly from random posts without first resorting to heuristics.





\bibliographystyle{ieeetr}
\bibliography{bibliography.bib}

\end{document}